\pdfoutput=1

\documentclass[11pt]{article}

\usepackage[]{ACL2023}

\usepackage{times}
\usepackage{latexsym}

\usepackage{times}
\usepackage{latexsym}
\usepackage{soul}
\usepackage{url}
\usepackage{graphicx}
\usepackage{amsmath}
\usepackage{multirow}
\usepackage{booktabs}
\usepackage{amsfonts}
\usepackage{comment}

\usepackage[T1]{fontenc}

\usepackage[utf8]{inputenc}

\usepackage{microtype}

\usepackage{inconsolata}

\title{Neural Machine Translation with Dynamic Graph Convolutional Decoder}
\author{\text{Lei Li}\thanks{Department of Computer Science and Technology, Tsinghua University, Beijing, China}, \text{Kai Fan}\thanks{Alibaba DAMO Academy, Alibaba Group Inc.}, \text{Lingyu Yang}\thanks{Tsinghua Shenzhen International Graduate School}, \text{Hongjia Li}\thanks{Tsinghua Shenzhen International Graduate School}, \text{Chun Yuan}\thanks{Tsinghua Shenzhen International Graduate School, Peng Cheng Lab} }
\begin{document}

\maketitle

\begin{abstract}
Existing wisdom demonstrates the significance of syntactic knowledge for the improvement of neural machine translation models.
However, most previous works merely focus on leveraging the source syntax in the well-known encoder-decoder framework. 
In sharp contrast, this paper proposes an end-to-end translation architecture from the (graph \& sequence) structural inputs to the (graph \& sequence) outputs, where the target translation and its corresponding syntactic graph are jointly modeled and generated. 
We propose a customized Dynamic Spatial-Temporal Graph Convolutional Decoder (Dyn-STGCD), which is designed for consuming source feature representations and their syntactic graph, and auto-regressively generating the target syntactic graph and tokens simultaneously. 
We conduct extensive experiments on five widely acknowledged translation benchmarks, verifying that our proposal achieves consistent improvements over baselines and other syntax-aware variants.
\end{abstract}

\begin{figure*}
	\centering
	\includegraphics[width=.99\textwidth]{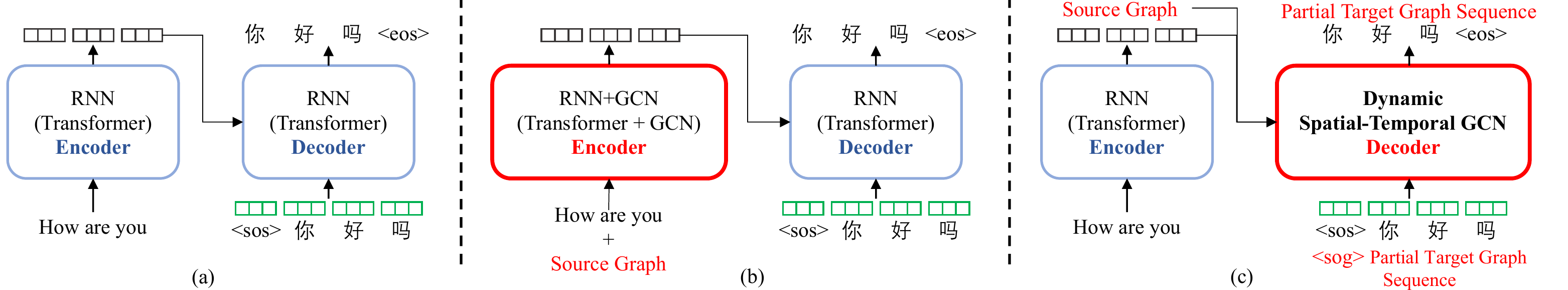}
	\caption{(a) The standard Seq2Seq translation models, (b) the syntax-aware Graph2Seq translation models, and (c) our introduced Dyn-STGCD to dynamically construct the syntactic graph and guide the translation simultaneously.}
	\label{fig1}
\end{figure*}

\section{Introduction}
\label{sec:intro}

Due to the success of deep learning in natural language processing, neural machine translation has made remarkable progress in recent years. 
Previous works such as statistical models \cite{wu2018dependency}, recurrent models \cite{wu2016google}, convolutional models \cite{gehring2017convolutional}, self-attention models \cite{vaswani2017attention}, and graph models \cite{bastings2017graph} have explored various methods to learn the context representation of tokens for the translation. 
Significantly, the Transformer \cite{vaswani2017attention}, only relying on multi-head attention networks, has achieved state-of-the-art performance across translation benchmarks.
Although these works have shown that they can learn some linguistic phenomena without explicit linguistic supervision, informing word representations with linguistic structures \cite{marcheggiani2018exploiting} can provide a useful signal for performance promotion.

Most previous works incorporate structured information into translation models on the source side. 
And syntax-aware NMT has gradually become a relatively prevalent topic. 
For example, \citet{nadejde2017predicting,bastings2017graph,wu2018dependency,stern2019insertion} claim improvements benefiting from treebank syntax. 
Because the syntactic information of the target sentence is unknown during inference, they typically build dependency trees based on recurrent models to represent syntactic relations of the source tokens and used them to implicitly guide the token generations in the decoder. 
Similarly, the recent work \cite{yang2021graphformers} only modifies the transformer encoder with Graph Neural Networks (GNN). 
We provide more related work discussions in Appendix \ref{app:related}.

We hypothesize two potential reasons why existing methods focus on exploiting the structure knowledge of the source side and modeling the implicit dependencies of the target sentence, rather than modeling the explicit syntax or any other linguistic structures on the target side. 
\textbf{(1)} It is technically difficult for adopting current popular auto-regressive architectures (Recurrent Neural Networks (RNN) and Transformer) to incorporate linguistic structures in the decoder. 
\textbf{(2)} It is ineffective to directly use conventional Graph Convolutional Networks (GCN) \cite{kipf2016semi} to model the auto-regressive property when generating the structure given partial target tokens. 
Therefore, we aim to close this gap by demonstrating how syntactic structures of target sentences can be explicitly modeled and partially generated in an auto-regressive manner. 
Our proposal can be differentiated from representative approaches and visualized in Figure \ref{fig1}.

Our main goal is to make the decoder accessible to rich syntactic structures and allow itself to decide which aspects of syntax are beneficial for the translation. 
We treat each token as a graph node and employ syntactic dependency or any available graph structural data as the relation (edge) indicator to construct both encoder and decoder syntactic graphs. 
In our work, the source syntactic graph is fixed and modeled as context information, while the target syntactic graph in the decoder is dynamically changed by inserting new node (the latest predicted token) and edges at each decoding step. 

Our main contributions are: 

\noindent
\textbf{(1)} We present a (graph \& sequence)-to-(graph \& sequence) model to incorporate syntactic structure, by jointly modeling and constructing the target translation and the corresponding syntactic graph. 

\noindent
\textbf{(2)} To fulfill the objective, we propose a novel auto-regressive graph decoder stacked by the dynamic spatial and temporal convolutional layers. 
It is compatible as general text generation task. 

\noindent
\textbf{(3)} Empirically, our approach can boost the performance on multiple translation benchmarks, significantly surpassing baselines and other syntax-aware variants, and providing additional outputs of target syntactic graphs. 
Our code will be released upon acceptance.

\section{Background and Notations}

\noindent 
\textbf{Machine Translation Model.}
Let $\mathbf{x} = {x_1, ..., x_U}$ and $\mathbf{y}  = {y_1, ..., y_T}$ be the source and the target sentences. 
Standard machine translation models the probability of $\mathbf{y}$ conditioned on serial data $\mathbf{x}$:
%
\begin{equation}\label{eq:mt}
P(\mathbf{y} |\mathbf{x} ;\theta )= \prod_{t=1}^{T} P(y_t|\mathbf{y} _{< t},\mathbf{x} ;\theta ).
\end{equation}
Existing translation models are generally equipped with an encoder-decoder structure and a target-to-source attention mechanism. 
The encoder encodes the source context, and the decoder generates the target tokens by attending the source context and performs left-to-right auto-regressive decoding.
In this work, we use transformer encoder layers \cite{vaswani2017attention} to extract the token embeddings of the source sentence, on top of which we build our syntax-aware decoder. 


\noindent \textbf{Graph Convolutional Networks.}
Given a static graph $G = (V,E)$ where $V$ is the set of vertices and $E$ is the set of edges. 
Its corresponding adjacency matrix is denoted as $A \in \mathcal{R}^{N\times N}$, where $A_{i, j}$ = 1 indicates $v_i, v_j \in V$ and $(v_i, v_j) \in E$ otherwise $A_{i, j}$ = 0.
GCN operates directly on graphs and aggregates information from immediate ($1$-hop) neighbors of nodes, and the information between longer-distance nodes is covered by stacking graph layers. 
At the $l$-th layer of GCN, it takes as input the node embeddings $H^{l-1}$ of the ($l$-$1$)-th layer and the adjacency matrix $A$, then output the updated embeddings $H^{l}$. 
The operation of one GCN layer \cite{kipf2016semi} $H^l = \text{GCN}(H^{l-1}, A, W^l)$ can be formulated as:
%
\begin{equation}\label{eq:gcn}
H^l = \sigma (\hat{D}^{-1/2}\hat{A}\hat{D}^{-1/2}H^{l-1}W^l),
\end{equation}
%
where $\hat{A} = A + I$ and diagonal degree matrix $\hat{D}_{ii} =\sum_j \hat{A}_{ij}$, 
$W^l$ is the layer specific trainable weights, and $\sigma$ is the activation function (e.g. ReLU).
Standard GCN \cite{kipf2016semi} and its variants \cite{hu2020heterogeneous,feng2022grand+,hou2022graphmae} fit well for fixed graph, however, we will extend it to a time-dependent variant that can model the dynamic graph sequence. 

\noindent \textbf{Graph Random Walks.}
Given two Graphs $G$ and $G^\prime$, their direct product $G_{\times}$ is a graph with vertex set $V_{\times} = \{(v, v^\prime)|v\in V \wedge v^\prime\in V^\prime\}$ and edge set $E_{\times} = \{(u,u^\prime), (v, v^\prime)|(u,v)\in E \wedge (u^\prime,v^\prime)\in E^\prime\}$. 
In the graph random walks \cite{nikolentzos2020random}, it is required to perform a direct product between the input graph $G$ and a ``hidden graph” $G^\prime$. 
The ``hidden graph” has trainable adjacency matrix and node embeddings, and it is analogous to trainable kernel weights in convolution.
Considering the node representations $X$ for input graph $G$ and trainable node embeddings $X^\prime$ for ``hidden graph” $G^\prime$, we define the $P$-step ($P\in\mathbb{N}$) random walk kernel which calculates the number of common walks of length $p$ between two graphs:
\begin{equation}\label{eq:grw}
\kappa^{(p)} \left ( G, G^\prime  \right ) = s^\top \left[A_{\times}^{p}\right] s, \text{ } \text{ } \text{ } \forall \text{ } 0\le p\le P 
\end{equation}
where $s = X^\prime X^\top$, and $A_{\times}$ is the adjacency matrix of $G_{\times}$ whose superscript $p$ indicates matrix power exponent.

\begin{figure}[t]
	\centering
	\includegraphics[width=.48\textwidth]{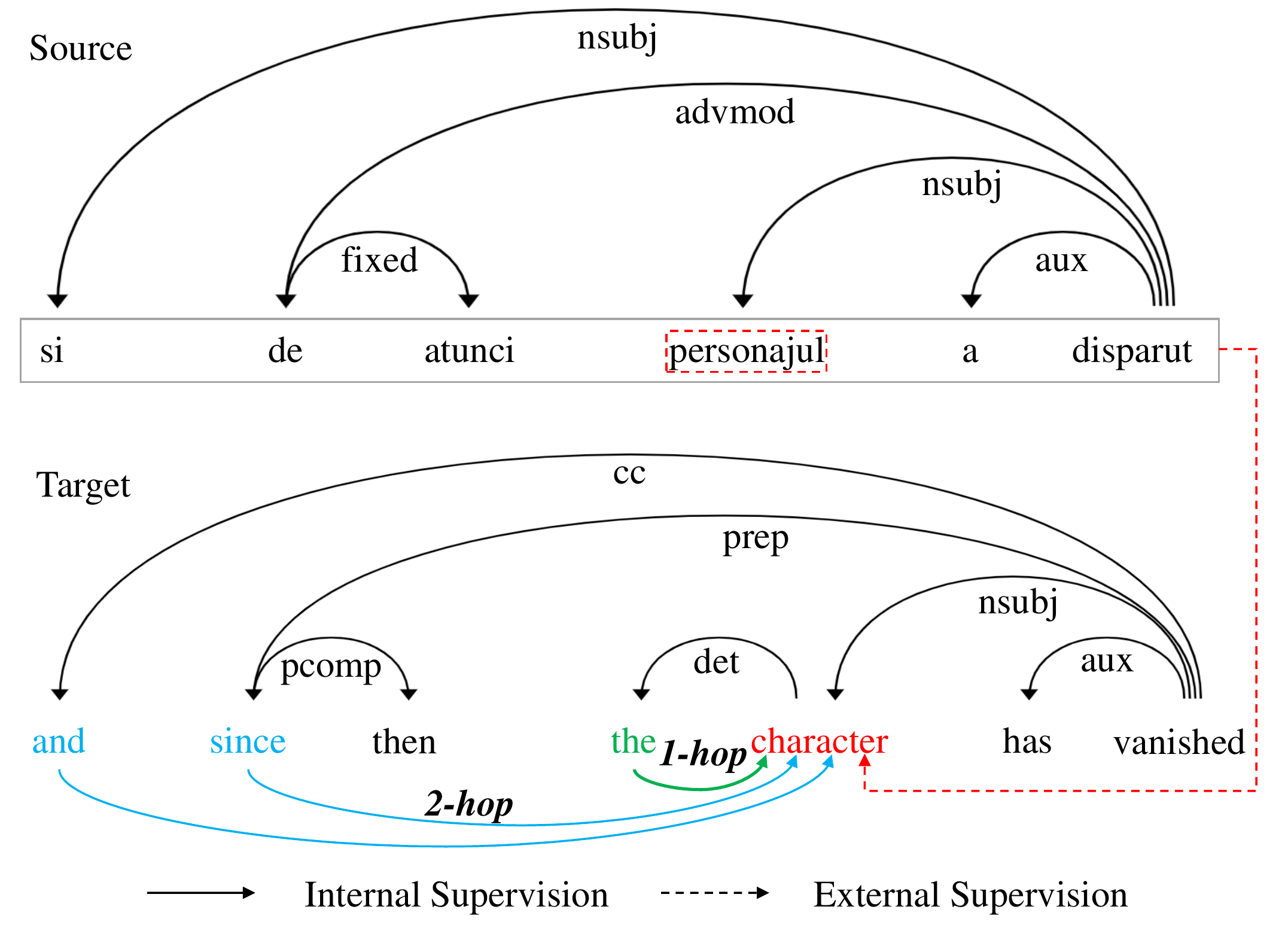}
	\caption{An example to show that tokens in different languages may partially share similar dependencies within the same semantics. To incorporate structure constraints, we leverage both internal and external supervision for constructing the syntactic graph in the decoder and guiding the translation simultaneously.}
	\label{fig6}
\end{figure}

\begin{figure*}
	\centering
	\includegraphics[width=.98\textwidth]{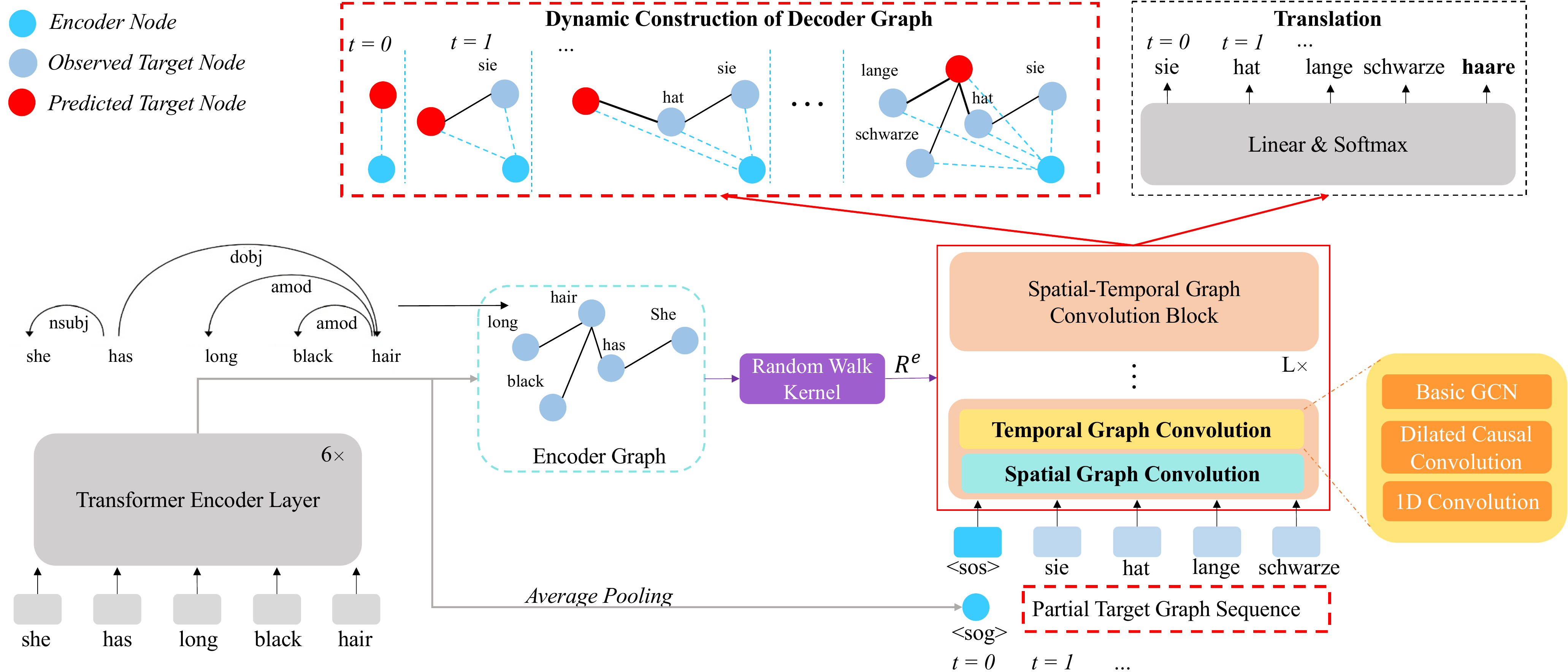}
	\caption{The overall architecture of our proposed framework. The partial target graph sequences in two red dashed boxes are the same, except that the time stamp of the decoder input is shifted by $+1$.}
	\label{fig:model}
\end{figure*}

\section{Our Approach} 
\label{approach}

\subsection{Motivation and Intuition}
Our work is inspired by the fact that tokens in different languages may partially share similar dependencies within the same semantics, and we give a persuasive example in Figure~\ref{fig6}. 
Regardless of the direction and the dependency type, the dependencies between tokens with the same semantics in the Romanian-English pair are remarkably similar. 
For instance, the syntactic dependency between `personajul' and `disparut' in the source language matches the corresponding dependency between `character' and `vanished' in the target language.
To incorporate structure information, we leverage both internal and external supervision to construct the syntactic graph of the decoder.

For internal supervision, we introduce GCN with dynamics to fit the (graph \& sequence)-to-(graph \& sequence) framework.
Specifically, we dynamically build the edges according to the predicted dependencies between each target node to form the syntactic graph.
In this case, when predicting the target token `character', the previously predicted token `the' is a 1-hop neighborhood, and the tokens `and' and `since' are 2-hop neighborhoods of `character' in the built syntactic graph.
Thus we can use the structure constraints inside the current syntactic graph to guide the generation of `character'.

For external supervision, we take advantage of the fact that graph random walks \cite{nikolentzos2020random} can effectively capture the structural similarity between two graphs, such that the source syntactic graph can be sustainably utilized to guide the translation outputs (target sentence \& its syntactic graph).
According to the similar syntactic dependencies between the source and target sentences, we perform the graph random walks on the syntactic graph of the source sentence to obtain external constraints, which are supplemented to the target token embeddings for jointly determining the decoder outputs.

\subsection{Overview of the Proposed Dyn-STGCD}
We model the joint distribution of the incrementally predicted target tokens at decoding and the dynamically constructed syntactic graph of the target sentence by factorizing it into the product of a series of conditional distributions:
\begin{equation}\label{eq:prob}
\begin{aligned}
    P(\mathbf{y}, \mathbf{y}^s|\mathbf{x}^s, \mathbf{x}) &= \prod_{t=1}^{T} P(y _t|\mathbf{y}^s_{\leq t},\mathbf{x} ^s,\mathbf{y}_{< t},\mathbf{x})  \\
   & \times P(y^s_t|\mathbf{y}^s_{< t}, \mathbf{x}^s, \mathbf{y}_{< t}, \mathbf{x})
\end{aligned}
\end{equation}
where $\mathbf{y}_{<t}, \mathbf{y}^s_{<t}$ are the partially generated target sentence and the corresponding syntactic graph, and $\mathbf{x}, \mathbf{x}^s$ indicate the content and structure knowledge of the entire source sentence.

Figure \ref{fig:model} shows the overall architecture of our proposal. 
Our approach first builds the source syntactic graph $\mathbf{x}^s$ with the output representations $X^e$ from the Transformer encoder, \emph{i.e.}, the output of the last encoder layer. 
In particular, the adjacency matrix $A^e$ is generated from the dependencies of source tokens, by pre-processing the source sentence with a parsing toolkit. 

For simplicity, a global average pooling is applied to $X^e$ to obtain the representative embeddings $\bar{X^e}$ of the source graph. 
We denote it as \texttt{<sog>} (start of graph) and keep it constant in the graph convolutions of the decoder, in other words, it literally plays the similar role of the special token \texttt{<sos>} (start of sentence) in the translation task. 

Meanwhile, we initialize the target adjacency matrix $A^{t=0}_{l=0}$ being diagonal, indicating only self-connections for a maximum number of candidate tokens(nodes) to be predicted in the decoder. 
In addition, all candidate nodes have an edge connection with the `Encoder Node' \texttt{<sog>}. 
Then we will introduce a Dynamic Spatial-Temporal Graph Convolutional Decoder (Dyn-STGCD) to auto-regressively update the syntactic graph.

As shown in Figure~\ref{fig:model}, Dyn-STGCD consists of stacked blocks of the spatial and temporal convolutions in an interleaving manner. 
In each block, we first use a spatial convolution layer where the GCN module is updated by a GRU cell and a neighbor-refined cell to perform the context-based self-adaption graph dynamism. 
At the last layer of the encoder and decoder, we conduct graph random walks on both the entire source syntactic graph and the partially constructed target syntactic graph, incorporating structure constraints to guide the translation. 
Then we adopt a temporal convolution layer that aggregates information from historical steps. 
Eventually, a simple linear and softmax layer will be added for target token prediction. 



\subsection{Spatial Convolution Layer}

\begin{figure*}
	\centering
	\includegraphics[height=.32\textwidth]{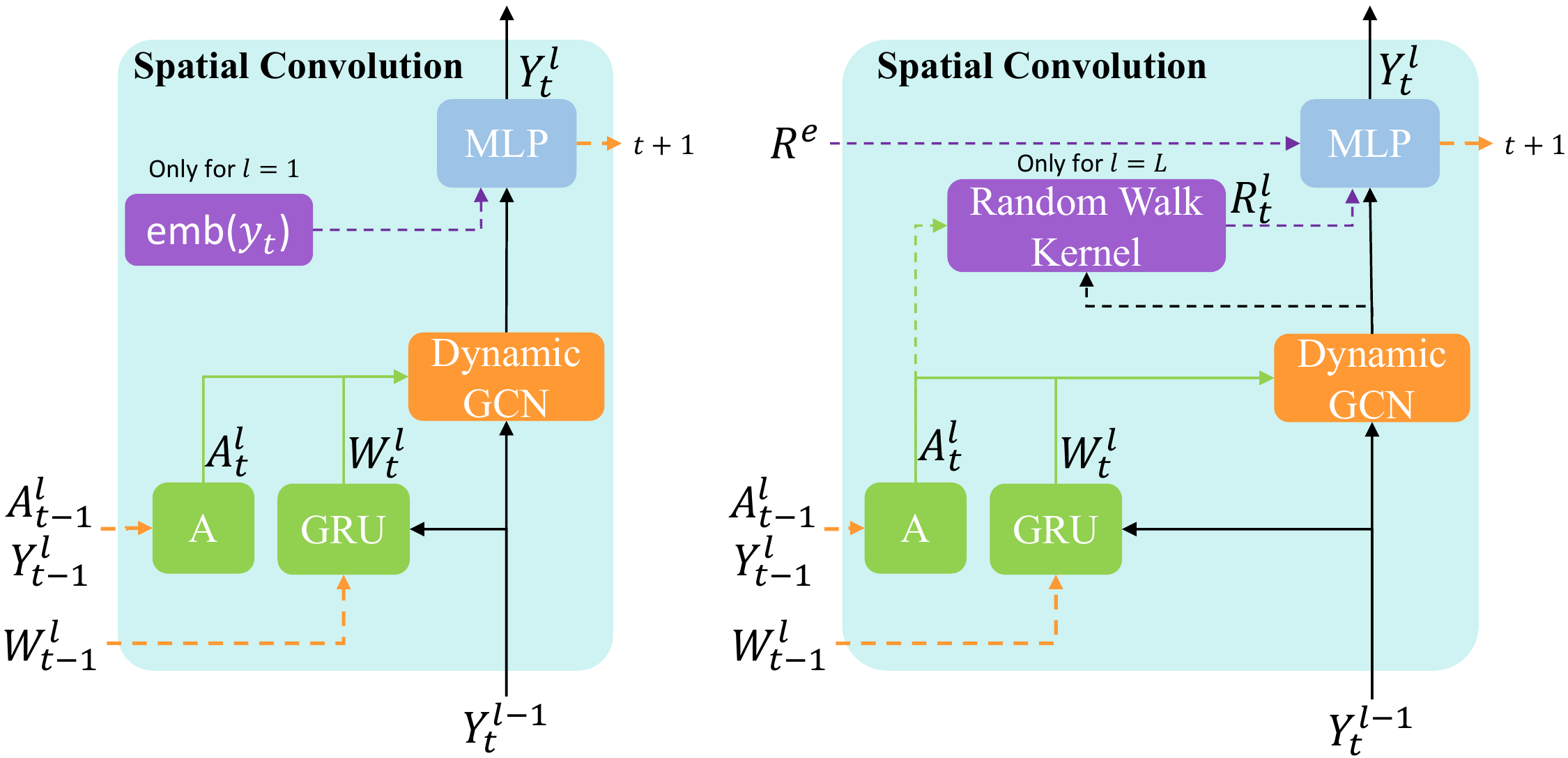}
	\includegraphics[height=.32\textwidth]{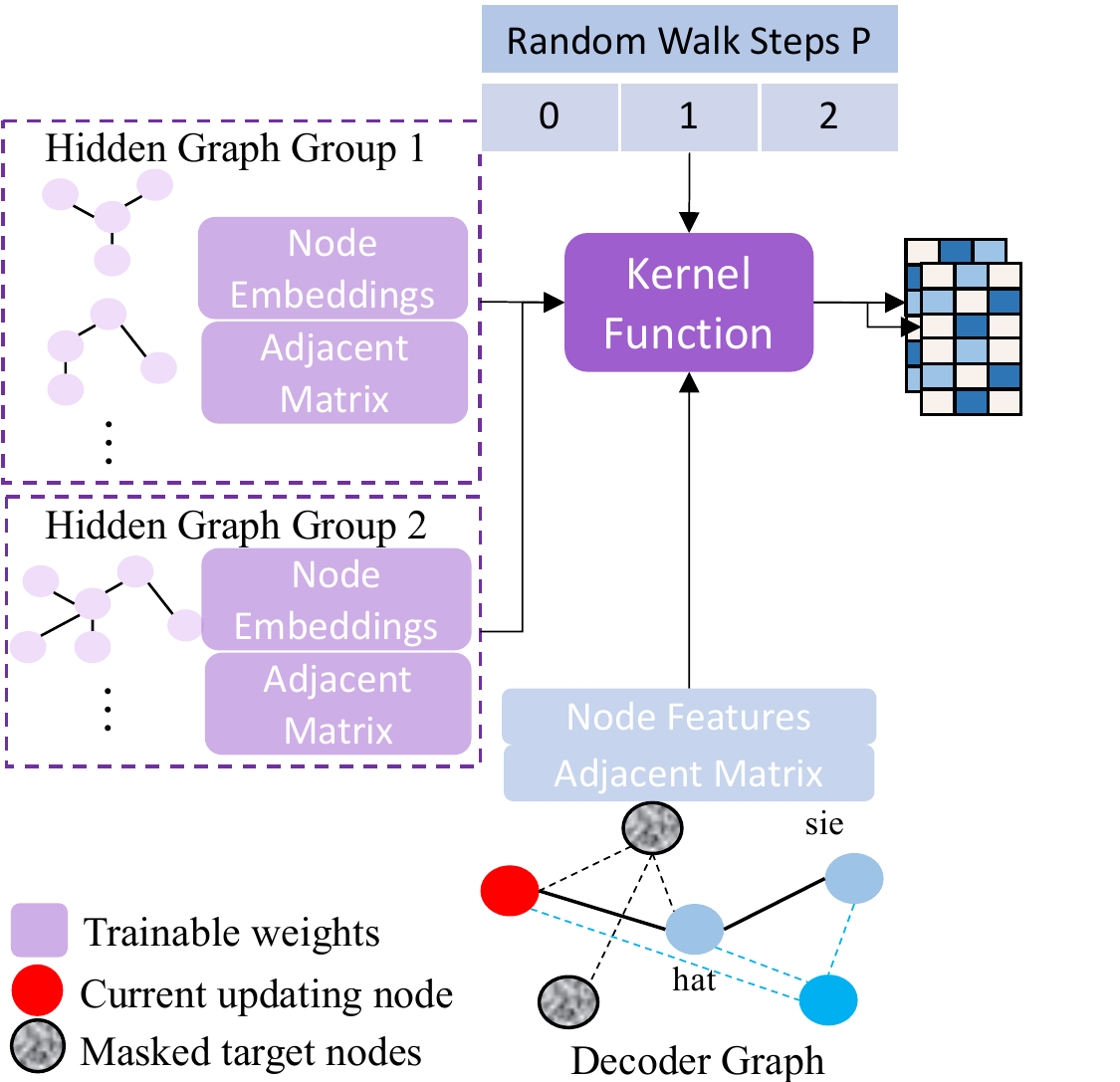}
	\caption{(Left and Middle): Spatial convolution module for different layers. (Right): Random walk kernel.}
	\label{fig:sc_rw}
\end{figure*}

\noindent \textbf{Dynamic GCN.}
We employ the GCN module to aggregate the neighborhood context of every snapshot in dynamic graphs. 
In order to extend the GCN with graph dynamism, we re-design the operation of the GCN layer in Eq.~(\ref{eq:gcn}) with subscript $t$, 
\begin{equation}\label{eq:dqcn}
Y^l_t = \text{DynamicGCN} (Y^{l-1}_t, A^l_t, W^l_t),    
\end{equation}
where $Y_t^l$ denotes the $l$-th layer graph features of target token at timestamp $t$ and $Y_t^0$ is initialized as $\mathbf{0}$ embedding.
Consider that the contributions of each neighborhood at different timestamps should be various, unlike the standard GCN, we propose the \textit{dynamic GCN} with the recurrent property for its weight matrix and adjacency matrix to dynamically control the impact of the context. 
Specifically, a recurrent architecture is adopted to update the weight matrix, and a neighbor-refined cell is used for optimizing the adjacency matrix composed of the current predicted target tokens. 
For weight matrix $W^l_t$ and adjacency matrix $A^l_t$ of $l$-th layer at timestamp $t$, we propose to use a GRU cell and a neighbor-refined cell, respectively.
\begin{align}
    W^l_{t} &= \text{GRU}(Y_t^{l-1}, W^l_{t-1}),  \label{eq:w_update} \\
    A^l_t &= \text{Sigmoid} \left (\text{Linear}\left ( A^l_{t-1} + \tilde{A}^l_{t-1} \right )\right ), \label{eq:a_update}
\end{align}
where $\tilde{A}^l_{t-1}$ defines raw adjacency matrix calculated by the graph attention coefficients \cite{velivckovic2017graph} between two observed nodes.
\begin{equation*}
     (\tilde{a}^l_{t-1})_{i, j} = \frac{e^{\text{LR}\left (\psi^{\top} \left [ W Y_{(t-1)i}^l \| W Y_{(t-1)j}^l \right ]   \right ) }}{\sum_{k\in \mathcal{N}_i }^{}e^{\text{LR}\left (\psi^{\top} \left [ W Y_{(t-1)i}^l \| W Y_{(t-1)k}^l \right ]   \right ) }}.
\end{equation*}
where $\psi$ and $W$ are trainable parameters, $\mathcal{N}_i$ denotes the neighbor set of node $i$, LR indicates the LeakyReLU activation function, and $\|$ means the row-wise concatenation.
Note that the graph attention module also takes as input an upper triangle causal masking matrix to prevent seeing future information. 
The left and middle pannels in Figure~\ref{fig:sc_rw} illustrate the detailed data flow through our proposed dynamic GCN. 

\noindent \textbf{Dynamic GCN with Random Walk.}
The graph random walk approach \cite{nikolentzos2020random} is feasible and effective in machine translation systems based on two essential properties: (1) the adopted ``hidden graphs” can learn the graph structures during training with back-propagation so that the translation outputs are highly interpretable, and (2) the employed random walk kernel is differentiable and therefore the whole framework is end-to-end trainable. 

For the Dynamic GCN in the last layer of the decoder, we add an extra module -- random walk kernel (see the right panel in Figure~{\ref{fig:sc_rw}}). 
In practice, we initialize two groups of trainable ``hidden graphs” with differentiated hyper-parameters. 
They will capture both the global and local structure representations of the target sentence by the kernel function in the random walk. 
One group includes 6 ``hidden graphs" with 6 nodes to capture global structure, and the other has 6 ``hidden graphs" with 4 nodes to capture local structure. 
The 6-node hidden graph can theoretically capture at most 5-hop structure via the random walk, approximately covering the longest distance in most syntactic graph. 
It is desired to learning global information. 

Concretely, given the maximum steps $P$ for graph random walk and $K$ ``hidden graphs” $\mathcal{G}^h =\left\{G^h_1, G^h_2, ..., G^h_K\right\}$, we can compute the graph random walk representations $R^l_t \in \mathcal{R} ^{K\times (P+1)}$ for graph $G^l_t$ at time step $t$ of $l$-th layer in the decoder $(R^l_t)_{kp} = \kappa^{(p)} \left( G_t^{l}, G_k^h\right)$, 
%
%
where the graph in the kernel argument is represented by its adjacency matrix and node representations. 
The detailed calculation can refer to the Appendix \ref{app:rw}. 
Recall we have two groups of hidden graphs, we slightly abuse notation $R_t^l$ to denote the concatenation of two resulted graph representations.

Eventually, spatial convolution will update the target representations as follows.
\begin{equation}\label{eq:dgcn_update}
    Y^l_t \leftarrow \left\{\begin{array}{lc}
    \text{MLP} (Y^l_t \| \text{emb}(y_t)),    \text{ }\text{ } \text{ } \text{ } l=1 \\
    \text{MLP} (Y^l_t),   \text{ }\text{ }\text{ } \text{ }  1<l<L \\
    \text{MLP} (Y^l_t \| R^e \| R^l_t),    \text{ } \text{ }\text{ } \text{ } l=L
    \end{array} \right. 
\end{equation}
%
where $R^e$ is the graph random walk representations of the source syntactic structure, MLP indicates the Multilayer Perceptron, emb(.) extracts the corresponding embeddings and $\|$ means the row-wise concatenation.


\subsection{Temporal Convolution Layer}

The representations of target nodes in the $l$-th graph layer are the aggregations of neighborhoods information. 
To capture temporal information in auto-regressive manner, we simply adopt the prevalent temporal convolution layer employing a TCN architecture \cite{bai2018empirical}. 
It stacks a 1-D fully convolution unit, a dilated causal convolution unit, and a basic graph convolution unit (Please refer to Appendix \ref{app:tcn} or reference for details).
This process is efficient for providing parallelism and aggregating rich historical information along the time axis. 
Besides, it requires less memory compared to RNN-based methods.
Therefore, we further use a basic GCN layer in Eq.~(\ref{eq:gcn}) to get the representations of each target token with current and historical information for both itself and its neighborhoods.

\subsection{Training Objective}

We optimize both the cross-entropy loss for the translation task and the smooth $L_1$ loss for the target graph prediction task. 
\begin{equation}\label{eq:loss}
    \mathcal{L} = \mathcal{L}_{CE} + \sum_{t=1}^{T} {L1}^{\text{smooth}}\left(A^L_t - A_t^{*}\right),
\end{equation}
where $A_t^{*}$ indicates the golden adjacency matrix of target syntactic graph at decoding step $t$. 
The smooth $L_1$ loss (see definition in Appendix \ref{app:l1s}) favors smooth and bounded gradient during back-propagation. 
It is worthwhile to mention that the training of graph sequences is free generation \cite{khorsi2019unsupervised} rather than teacher forcing \cite{lamb2016professor}, unless we substitute $A_{t-1}^l$ for $A_{t-1}^*$ in Eq.~(\ref{eq:a_update}). 
Following prior work \cite{stern2019insertion}, we also test our method whether the additional distribution-level knowledge distillation (KD) \cite{kim2016sequence} loss will help.

\section{Experiments}

\noindent \textbf{Datasets.} We evaluate on five widely acknowledged translation benchmarks: the NIST Chinese-English, the WAT 2016 Japanese-English, the WAT 2016 English-Japanese, the WMT 2017 Chinese-English and the WMT 2017 English-German translation tasks. 
To train our model, we need both sides’ dependency relations. 
As there are no golden annotations of dependency relations in all the training corpora, we use pseudo parsing results from parser tools (see details in Appendix \ref{app:dataset}). 

\noindent \textbf{Baselines.} We compare our proposal with several baselines including HPSMT \cite{chiang2005hierarchical}, SMT Hiero (S2T, T2S) \cite{wu2018dependency}, (6-layer) RNNearch \cite{bahdanau2014neural}, and vanilla Transformer \cite{vaswani2017attention}. 
In addition, most existing syntax-aware translation methods that use source consistency/dependency trees or target consistency/dependency trees are also included, including ANMT \cite{eriguchi2016character}, Tree2Seq \cite{chen2017improved}, Bpe2tree \cite{aharoni2017towards}, Seq2Dep \cite{le2017improving}, BiRNN+GCN \cite{bastings2017graph}, Semantic GCN \cite{marcheggiani2018exploiting}, and SE+SD-NMT \cite{wu2018dependency}.
More details are provided in the Appendix \ref{app:baselines}.  
Further, we compare with several latest works including HPT \cite{hassan2018achieving}, LightConv \cite{wu2019pay}, DynamicConv \cite{wu2019pay}, Admin \cite{liu2020understanding}, ReZero \cite{bachlechner2021rezero}, NormFormer \cite{shleifer2021normformer}
and DeepNet (6L-6L) \cite{wang2022deepnet} on WMT 2017.

\begin{table}[t]
\centering
\resizebox{0.48\textwidth}{!}{
\setlength{\tabcolsep}{0.4pt}
\begin{tabular}{l|c|c|c}
\toprule
\textbf{Method}  & \textbf{NIST2008} & \textbf{NIST2012} & \textbf{Average} \\ \toprule 
HPSMT                       & 26.1     & 27.5     & 26.8    \\ 
RNNearch                    & 31.6     & 29.0     & 30.3    \\ \hline 
Seq2Dep                      & 31.5     & 29.9     & 30.7    \\ 
Bpe2tree                     & 31.7     & 29.8     & 30.8    \\ \hline
Tree2Seq                     & 31.9     & 30.4     & 31.2    \\ 
BiRNN+GCN                   & 32.0     & 30.5     & 31.3    \\ \hline 
SD-NMT                        & 33.1     & 31.4     & 32.3    \\ 
SE+SD-NMT                    & 33.8     & 31.8     & 32.8    \\ \hline
Transformer                       & 37.5     & 37.3     & 37.4    \\ \bottomrule 
\textbf{Dyn-STGCD}          & \textbf{39.2}     & \textbf{37.9}     & \textbf{38.6}    \\ \bottomrule
\end{tabular}
}
\caption{Evaluation results on the NIST Chinese-English tasks with BLEU metric.}
\label{tab1}
\end{table}

\begin{table}[t]
\centering
\resizebox{0.48\textwidth}{!}{
\setlength{\tabcolsep}{0.4pt}
\begin{tabular}{l|cc|cc}
\toprule
\multirow{2}{*}{\textbf{Method}}       & \multicolumn{2}{c|}{\textbf{Ja-En}} & \multicolumn{2}{c}{\textbf{En-Ja}} \\ \cline{2-5} 
                              & \multicolumn{1}{c|}{BLEU}   & RIBES   & \multicolumn{1}{c|}{BLEU}   & RIBES   \\ \toprule
SMT S2T (T2S)                  & \multicolumn{1}{c|}{20.4}   & 67.8    & \multicolumn{1}{c|}{33.4}   & 75.8    \\ 
RNNearch                      & \multicolumn{1}{c|}{23.5}   & 74.6    & \multicolumn{1}{c|}{34.8}   & 80.9    \\ 
Transformer                      & \multicolumn{1}{c|}{25.6}   & 75.1    & \multicolumn{1}{c|}{35.7}   & 81.3    \\ \midrule
ANMT         & \multicolumn{1}{c|}{-}      & -       & \multicolumn{1}{c|}{34.9}   & 81.7    \\ 
Bpe2tree                       & \multicolumn{1}{c|}{24.4}   & 74.8    & \multicolumn{1}{c|}{-}      & -       \\ 
Seq2Dep                        & \multicolumn{1}{c|}{24.2}   & 74.7    & \multicolumn{1}{c|}{-}      & -       \\ 
SE+SD-NMT                       & \multicolumn{1}{c|}{26.3}   & 75.7    & \multicolumn{1}{c|}{36.4}   & 81.8    \\ \bottomrule
\textbf{Dyn-STGCD}                      & \multicolumn{1}{c|}{\textbf{27.1}}   & \textbf{75.9}    & \multicolumn{1}{c|}{\textbf{38.1}}   & \textbf{82.6}    \\ \bottomrule
\end{tabular}
}
\caption{Evaluation results on the WAT 2016 translation task with BLEU and RIBES.}
\label{tab2}
\end{table}

\begin{table}
\centering
\resizebox{0.48\textwidth}{!}{
\setlength{\tabcolsep}{0.4pt}
\begin{tabular}{l|c|l|c}
\toprule
\textbf{Method}            & \textbf{Zh-En} & \textbf{Method} & \textbf{En-De} \\ \toprule
Transformer  & 23.8   & ReZero  & 26.9     \\ 
HPT  & 24.2  & Admin   & \textbf{27.9}    \\ 
LightConv   & 24.3  & NormFormer    & 27.0     \\ 
DynamicConv   & 24.4  & DeepNet (6L-6L)  & 27.8     \\ 
\bottomrule
\textbf{Dyn-STGCD}  & \textbf{24.6}  & \textbf{Dyn-STGCD}   & 27.6     \\ \bottomrule
\end{tabular}
}
\caption{Evaluation results on the WMT 2017 benchmark with BLEU. Noted that previous GNN or syntactic graph related methods are mainly based on RNN architectures, while the methods compared are recent novel translation architectures which are not necessarily syntactic graph related. It shows our method achieves competitive BLEU with other novel architectures.}
\label{tab3}
\end{table}


\noindent \textbf{Implementation Details.} 
For model implementation, we implement our proposed dynamic GCN based on the orginal GCN in the GNN library DGL\footnote{\scriptsize{\url{https://www.dgl.ai/}}} and the random walk kernel with graph random walk library\footnote{\scriptsize{\url{https://github.com/giannisnik/rwgnn}}}. 
It is convenient to integrate our proposed modules into transformer model.

During training, we set the source and target vocabulary sizes to 30K. 
The encoder has 6 layers of attention with 4 attention heads each, with the embedding size and hidden states 512, and the feed-forward layer hidden size 2048. 
The decoder contains a 4-block Dyn-STGCD, whose spatial convolution layers perform $\mathcal{P} =\left \{ 0, 1\right \} $-steps random walk. 
We use 8 NVIDIA GeForce RTX 2080 Tis to conduct all experiments. 

During inference of the graph generation, we assign a new empty node at the beginning of each step, then calculate the edge information with previously generated nodes, and eventually, we predict the text token and update the empty node. 
Beam search is only applied for text token prediction.
The evaluation results are reported with the case-sensitive BLEU\footnote{\scriptsize{\url{https://github.com/awslabs/sockeye/tree/master/contrib/sacrebleu}}}. 
For the Ja$\leftrightarrow$En tasks, we use the official evaluation procedure provided by WAT 2016\footnote{\scriptsize{\url{http://lotus.kuee.kyoto-u.ac.jp/WAT/evaluation/index.html}}}, where the suggested RIBES \cite{isozaki2010automatic} is also used for evaluation. 
For fair comparison with all previous works, we do not apply the trick such as averaging the last several checkpoints.

\subsection{Text Generation}

\noindent \textbf{Quantitative Results.} Table \ref{tab1} shows the evaluation results over two test sets on the NIST Zh-En task. 
We can see that the RNN based syntax-based models outperform the RNNearch baseline, indicating syntax is helpful for translation. 
In terms of the source side, compared to previous source syntax-aware methods (Tree2Seq and BiRNN+GCN), our method achieves significantly better average BLEU scores, showing that target dependency structures with syntactic context are helpful for translation.
In terms of the target side, our proposal achieves higher BLUE scores than SE+SD-NMT, which demonstrates that modeling the dependency structure of target sentences explicitly through graph networks is more effective than implicitly parsing through RNN. 
Table \ref{tab2} shows the evaluation results on the Japanese-English (Ja-En) and the English-Japanese (En-Ja) tasks.
Significantly, our method also achieves the highest scores in terms of BLEU and RIBES in both translation directions.

To further verify the effect of syntax knowledge modeled by our graph-based approach, we conduct the experiments on the Zh-En and the En-De tasks of the WMT 2017 large-scale benchmark. 
As shown in Table \ref{tab3}, our approach outperforms the vanilla Transformer and its variants including ReZero and NormFormer. 
It indicates although the Transformer is good at modeling long-distance relations, our method can incorporate rich syntactic knowledge to further improve the performance of translation. 
Besides, our Dyn-STGCD performs better than DynamicConv which is also based on dynamic convolutions, validating the powerful ability of graph networks to capture dynamic dependencies at the decoder.  

\begin{table}
\centering
\resizebox{0.48\textwidth}{!}{
\setlength{\tabcolsep}{0.4pt}
\begin{tabular}{lcccc}
\toprule
\multicolumn{1}{l|}{\textbf{Ablation Models}}     & \multicolumn{1}{c|}{\textbf{Ja-En}} & \multicolumn{1}{c|}{\textbf{Gain}} & \textbf{En-Ja}  & \multicolumn{1}{|c}{\textbf{Gain}} \\ \toprule
\multicolumn{1}{l|}{Transformer} & \multicolumn{1}{c|}{25.6}  &\multicolumn{1}{c|}{-1.5}                                &   35.7        & \multicolumn{1}{|c}{-2.4}                   \\ \hline
\multicolumn{5}{c}{\textbf{Dyn-STGCD}} \\ \toprule
\multicolumn{1}{l|}{\textbf{Baseline}}                      & \multicolumn{1}{c|}{\textbf{27.1}}       &\multicolumn{1}{c|}{---}                            &   \textbf{38.1}             &\multicolumn{1}{|c}{---}                \\ 
\multicolumn{1}{l|}{w/o $\mathcal{P}$}                              & \multicolumn{1}{c|}{26.0}        &\multicolumn{1}{c|}{-1.1}                          & 36.5          &\multicolumn{1}{|c}{-1.6}                       \\ 
\multicolumn{1}{l|}{w/o source of $\mathcal{P}$}                              & \multicolumn{1}{c|}{26.5}       &\multicolumn{1}{c|}{-0.6}                           & 37.5            &\multicolumn{1}{|c}{-0.6}                     \\ 
\multicolumn{1}{l|}{$\mathcal{P} =\left \{ 0, 1\right \} $ (baseline)}                              & \multicolumn{1}{c|}{27.1}      &\multicolumn{1}{c|}{---}                            & 38.1                &\multicolumn{1}{|c}{---}                \\ 
\multicolumn{1}{l|}{$\mathcal{P} =\left \{ 0, 1, 2\right \} $}                              & \multicolumn{1}{c|}{\textbf{27.5}}   &\multicolumn{1}{c|}{+0.4}                               &    38.2            &\multicolumn{1}{|c}{+0.1}                 \\ 
\multicolumn{1}{l|}{$\mathcal{P} =\left \{ 0, 1, 2, 3\right \} $}                              & \multicolumn{1}{c|}{27.2}     &\multicolumn{1}{c|}{+0.1}                             &  \textbf{38.4}                  &\multicolumn{1}{|c}{+0.3}             \\ \hline
\multicolumn{1}{l|}{w/o KD loss}                              & \multicolumn{1}{c|}{27.0}     &\multicolumn{1}{c|}{-0.1}                             &  37.8      &\multicolumn{1}{|c}{-0.3}                         \\  \bottomrule
\end{tabular}
}
\caption{Ablation studies on the adopted random walks and the special trick for translation.}
\label{tab4}
\end{table}

\begin{figure}
	\centering
	\includegraphics[width=.49\textwidth]{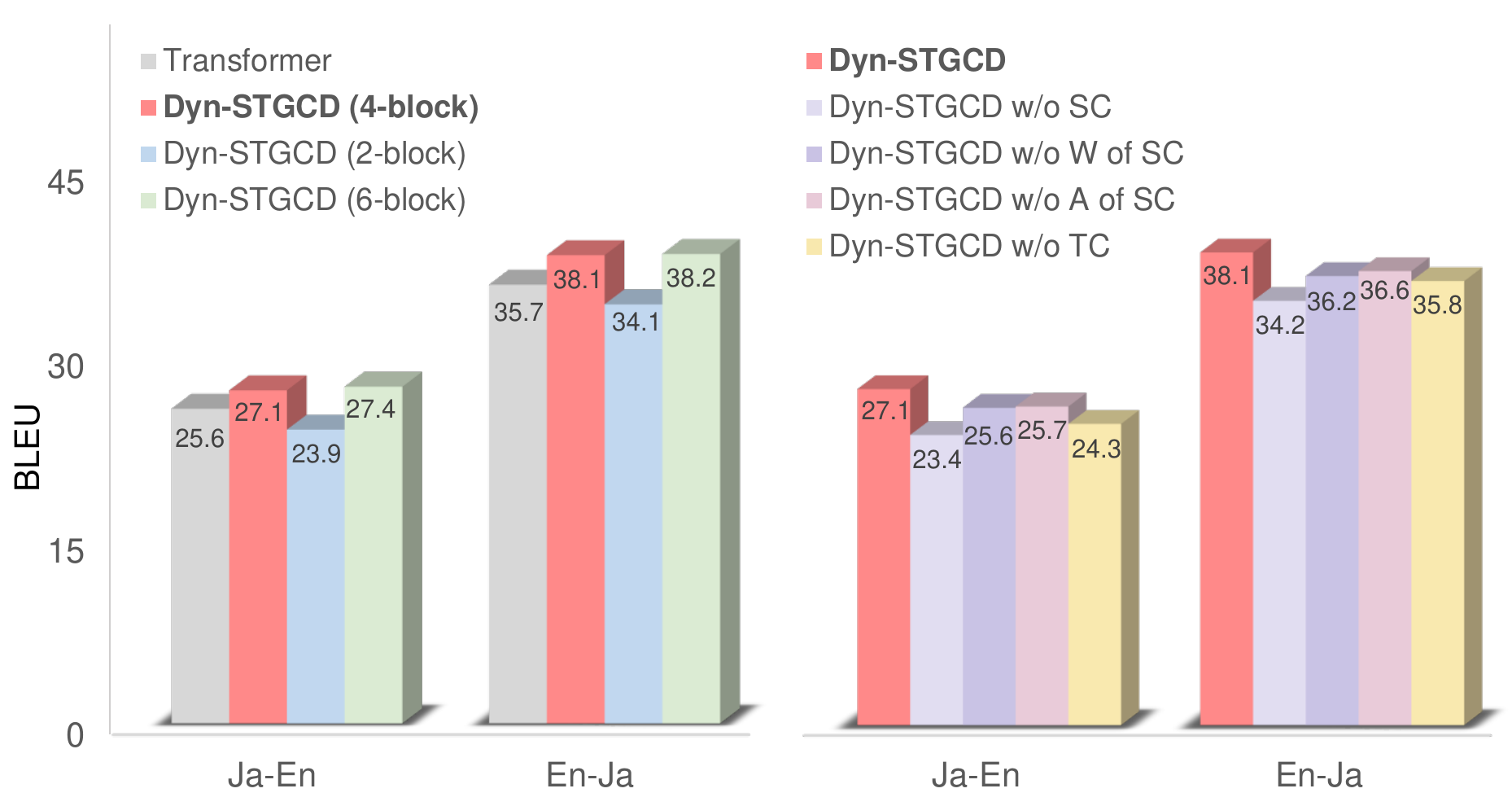}
	\caption{Ablation studies on the decoder length, spatial and temporal layers of the proposed Dyn-STGCD.}
	\label{fig15}
\end{figure}

\noindent \textbf{Ablation Analysis.} 
We conduct ablation studies of the proposed decoder on two datasets and visualize in Figure \ref{fig15}. 
When substituting the introduced spatial-temporal blocks for the transformer decoder, our method (with a 4-block decoder) can gain 1.5 and 2.4 BLEU points. 
However, as the blocks continue to be stacked (6-block), there is no significant gain but an apparent increase in model size. 
Besides, the introduced spatial convolution (SC) is essential for performance promotion (with an absolute improvement of 3.7 and 3.9 BLEU points), while the dynamic properties of both $W$ and $A$ in Eq.~(\ref{eq:w_update}) and (\ref{eq:a_update}) have a strong positive impact. 
Similarly, the temporal convolution (TC) layer has a positive contribution as well. 

To investigate the effectiveness of introduced graph random walks, we ablate the number of steps of graph random walks on the predicted target nodes in Table \ref{tab4}. 
Notably, the quality of translation decreases when we remove the random walks on the source tokens, suggesting that both source and target syntactic information benefit translation models, and their effects can be accumulated. 
This also shows that our framework not only expands the form of translation input, but also confirms that the input of additional syntactic graph structure effectively improves the performance. 
Besides, it can be inferred that increasing the step length (e.g., $\mathcal{P} =\left \{ 0, 1, 2\right \} $) can improve the capability of ``hidden graphs” to capture the syntactic structure. 
However, continuing to increase the step length (e.g., $\mathcal{P} =\left \{ 0, 1, 2, 3\right \} $) will not always improve the performance, while it not only introduces more parameters but also is likely to confuse the model by the complicated closed-loop structure in the graph. 

In addition, we observe only a slight BLEU drop without KD training, indicating that our method does not rely on KD.

\noindent \textbf{Inference Performance.} In Table~\ref{tab:infer}, we compare the inference speed performance of our base model with transformer base architecture. 
In general, our method is on par with the well known baseline, and indicates its applicability in real scenario. 

\noindent
\textbf{Additional Experiments.} More experiments on NIST Zh-En and WMT2014 En-De can refer to Table \ref{tab5} and Table \ref{tab7} in Appendix \ref{app:app_exp}, respectively.

\begin{figure}
	\centering
	\includegraphics[width=.48\textwidth]{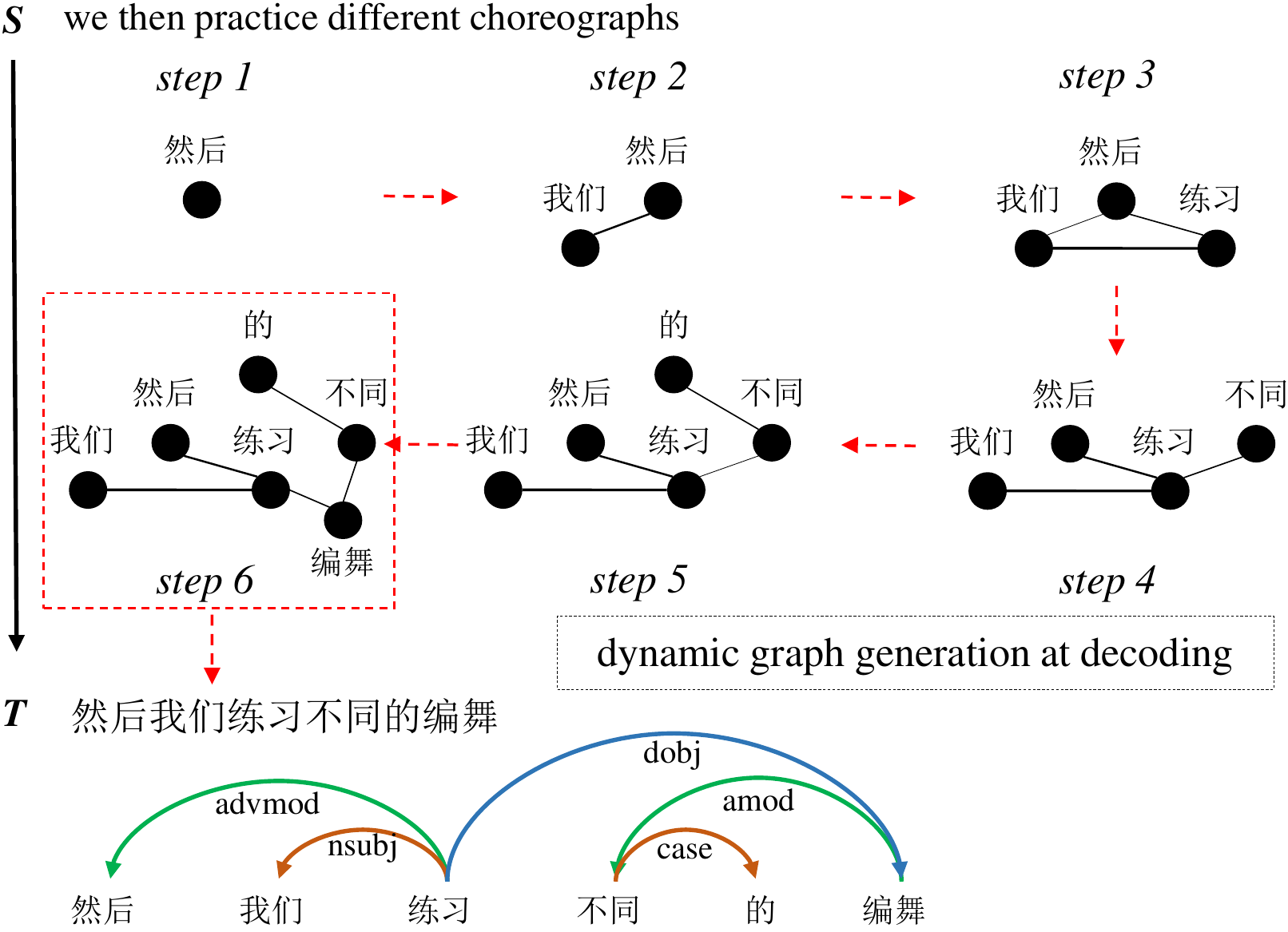}
	\caption{Dynamically construction of the syntactic graph by our proposed Dyn-STGCD for an En-Zh pair.}
	\label{fig5}
\end{figure}

\subsection{Graph Generation}

\noindent \textbf{Syntactic Graph Generation.} 
We visualize the dynamic graph generation process inferred by our Dyn-STGCD in Figure \ref{fig5}. 
As a by-product, it could explicitly reflect the dependencies between each target token.
Besides, we parse the target sentence by the pre-trained parsing tool to acquire the pseudo dependency reference, demonstrating that the generated syntactic graph is highly consistent with the parsing result. 
We also provide an approximate estimation of the syntactic graphs prediction and achieve an unlabeled attachment score \cite{plank2015dependency} (UAS) of \textbf{95.67}\% (see more details in the Appendix \ref{app:app_exp}), which demonstrates that the generated syntactic graphs are much similar to the parsing results from the stand-alone parser.
%

\begin{figure}[t]
	\centering
	\includegraphics[width=.48\textwidth]{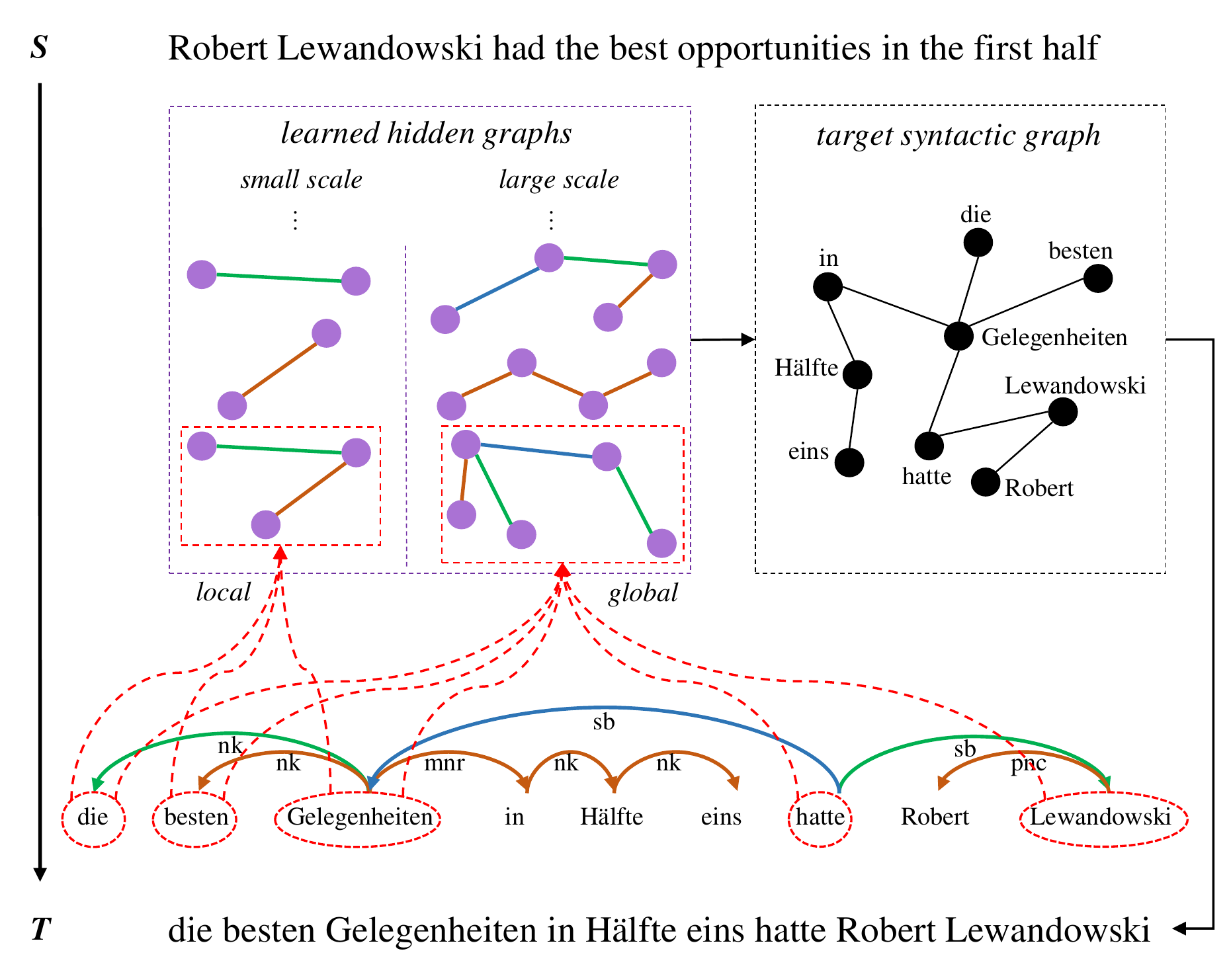}
	\caption{Visualizations of learned ``hidden graphs” and the constructed syntactic graph by Dyn-STGCD. $S$: Source sentence, $T$: Translation output. The ground-truth syntactic dependencies among target tokens parsed by the parsing-tool are given.}
	\label{fig7}
\end{figure}

\noindent \textbf{Qualitative Visualization.} In Figure \ref{fig7}, we present the visualizations of both the learned ``hidden graphs” and the constructed syntactic graph for an English-German pair.
We parse the target sentence by parsing tool to acquire the pseudo dependency reference, demonstrating that the constructed syntactic graph is highly consistent with the parsing result.
Significantly, it can be clearly observed that by introducing hierarchical graph random walks at the decoding stage, the ``hidden graphs” can capture both the local and global dependencies of target sentences, making the model obtain more syntax knowledge for guiding better translation.

\begin{table}
\centering
\resizebox{0.48\textwidth}{!}{
\setlength{\tabcolsep}{0.4pt}
\begin{tabular}{l|c|c|c}
\toprule
\textbf{Model} & \textbf{Arch.} &\textbf{\# params.} & \textbf{Speed} \\
\toprule
Transformer & 6L-6L & 36.8M & 52.3 toks/s \\
\bottomrule
\textbf{Dyn-STGCD} & \textbf{6L-4L} & \textbf{33.7M} & \textbf{57.1 toks/s} \\
\bottomrule
\end{tabular}
}
\caption{Inference performance comparison.}
\label{tab:infer}
\end{table}

\section{Conclusion}
We propose a novel decoder that can possibly replace conventional decoders (\emph{e.g.}, the RNN or transformer layers) in the encoder-decoder framework. 
Our approach can jointly model the generation of the text and related graph data. 
We verify our proposal on NMT model with syntactic knowledge, and the experimental results show that it is competitive compared to other related prevailing models. 
We expect our work could pioneer the research of structural generation for the graph decoders.

\bibliography{custom}
\bibliographystyle{acl_natbib}

\appendix



\begin{table*}
\centering
\begin{tabular}{cccccc}
\toprule
\textbf{Method}                  & \textbf{NIST2005} & \textbf{NIST2006} & \textbf{NIST2008} & \textbf{NIST2012} & \textbf{Average} \\ \toprule
HPSMT                    & 35.3     & 33.6     & 26.1     & 27.5     & 30.6    \\ 
RNNsearch                & 38.1     & 39.0     & 31.6     & 29.0     & 34.4    \\ \hline 
Seq2Dep                  & 38.9     & 39.0     & 31.5     & 29.9     & 34.8    \\ 
Bpe2tree                 & 39.0     & 39.3     & 31.7     & 29.8     & 34.9    \\ \hline 
Tree2Seq                 & 39.0     & 40.3     & 31.9     & 30.4     & 35.4    \\ 
BiRNN+GCN                & 39.0     & 41.1     & 32.0     & 30.5     & 35.7    \\ \hline 
SD-NMT                   & 39.4     & 41.8     & 33.1     & 31.4     & 36.4    \\ 
SE+SD-NMT                & 40.1     & 42.3     & 33.8     & 31.8     & 37.0    \\ \hline 
Transformer                   & 45.5     & \textbf{46.7}     & 37.5     & 37.3     & 41.8    \\ \toprule 
\textbf{Dyn-STGCD}               & \textbf{46.1}     & 46.5     & \textbf{39.2}     & \textbf{37.9}     & \textbf{42.4}    \\ \toprule
\end{tabular}
\caption{Evaluation results on the NIST Chinese-English tasks with BLEU metric.}
\label{tab5}
\end{table*}

\section{More Details of Our Approach}
\label{sec:appendix}

\subsection{Spatial Convolution Layer} 

The updating for weight matrix of the $l$-th layer at time $t$ is formulated as:
\begin{align}
    W^l_{t} = \text{GRU}(Y_t^{l-1}, W^l_{t-1}) \nonumber \\
    = (1-Z^l_t) \cdot W^l_{t-1} + Z^l_t \cdot \widetilde{W}^l_{t}, \nonumber
\end{align}
where
\begin{align}
    Z^l_t = \text{Sigmoid}(P_Z^lY_t^{l-1} + Q_Z^lW_{t-1}^{l} + B_Z^l), \nonumber\\
    R^l_t = \text{Sigmoid}(P_R^lY_t^{l-1} + Q_R^lW_{t-1}^{l} + B_R^l), \nonumber \\
    \widetilde{W}^l_{t} =  \text{Tanh}(P_W^lY_t^{l-1} + Q_W^l(R^l_t \cdot W_{t-1}^{l}) + B_W^l). \nonumber
\end{align}
We apply the standard GRU operation on each column of the involved matrices independently since standard GRU maps vectors to vectors but we have matrices here. 
We treat $W^l_{t-1}$ as the hidden state of GRU, and the embedding vector $Y_t^{l-1}$ is chosen as the input of GRU at every time step to represent current information.
$Z^l_t$, $R^l_t$, and $\widetilde{W}^l_{t}$ are the update gate output, the reset gate output, and the pre-output, respectively.
To deal with the discrepancy of the column size between weight matrix $W^l_{t-1}$ and embedding matrix $Y_t^{l-1}$, a summarization \cite{cangea2018towards} on $Y_t^{l-1}$ is further added to the evolving graph convolution layer to transfer $Y_t^{l-1}$ to have the same column size as $W^l_{t-1}$.

\subsection{Random Walk on Two Graphs} 
\label{app:rw}

Consider the currently built graph $G^l_t$ at time step $t$ in the decoder and a ``hidden graph" $G^h$ introduced in \cite{nikolentzos2020random} with trainable node embeddings $K^h$, their direct product $G^{l,h}_{t,\times}$ is a graph over pairs of vertices from $G^l_t$ and $G^h$, and two vertices in $G^{l,h}_{t,\times}$ are neighbors if and only if the corresponding vertices in $G^l_t$ and $G^h$ are both neighbors. 
It has been shown \cite{nikolentzos2020random} that performing a random walk on the direct product $G^{l,h}_{t,\times}$ is equivalent to performing a simultaneous random walk on the two graphs $G^l_t$ and $G^h$.

In this way, the random walk kernel will count all pairs of matching walks on $G^l_t$ and $G^h$ through the adjacency matrix of their direct product graph. 
Mathematically, the adjacent matrix $A^{l,h}_{t,\times}$ can be calculated as follows:
\begin{equation}
    A^{l,h}_{t,\times} = A_t^l \otimes A^h, \nonumber
\end{equation}
where $\otimes$ means Kronecker product.

Then we denote vector:
\begin{equation}
    s = {K^h \left(Y_t^l\right)^\top}, \nonumber
\end{equation}
where $Y_t^l$ is the target node representation.
Finally, we can calculate the kernel function:
\begin{gather}
    \kappa^{(p)} \left ( G^l_t, G^h  \right ) = \sum_{i=1}^{\left | V^{l,h}_{t,\times} \right | } \sum_{j=1}^{\left | V^{l,h}_{t,\times}  \right | } \left [ (ss^\top) \odot \left(A^{l,h}_{t,\times}\right)^p \right ]_{ij} \nonumber \\ = s^\top \left(A^{l,h}_{t,\times}\right)^p s, \nonumber
\end{gather}
where $\left(A^{l,h}_{t,\times}\right)^p$ means the matrix with power exponent $p$ and $\odot$ means the element-wise multiplication, and $V^{l,h}_{t,\times}$ means the vertex set of direct product graph. 

\subsection{Dilated Causal Convolution}
\label{app:tcn}
Formally, given a length-$M$ sequence of node representations $Y^l\in R^{T\times D}$ in the $l$-th layer with $D$ channels, and a filter $f : \{0, 1,...h-1\}$, the temporal convolution operation $H$ on $t$-th element of the sequence $Y^l$ is formalized as follow:
\begin{equation}
    H(t) = (Y^l\ast _{d} f)(t) = \sum_{i=0}^{h-1} f(i)\cdot Y^l_{t-d\cdot i}, \nonumber \noindent
\end{equation}
where $d$ is the dilation factor, $k$ is the filter size, and $t - d \cdot i$ indicates the direction of the past.

\subsection{Smooth $L_1$ Loss}
\label{app:l1s}
We use the following smooth $L_1$ loss to supervise the training graph adjacent matrix, which favors smooth and bounded gradient during back-propagation. 
To see this motivation clearly, we also provide the derivative of smooth $L_1$ loss function:
\begin{align}
    \text{smooth}_{L_1}(x) &= \left\{\begin{array}{cc}
        0.5x^2, & |x| \leq 1  \\
        |x| - 0.5. & \text{otherwise}
    \end{array} \right. \nonumber \\
    \frac{\partial \text{smooth}_{L_1}(x)}{\partial x} &= \left\{\begin{array}{cc}
        x, & |x| \leq 1  \\
        -1, & x < -1 \\
        1. & x > 1  \nonumber
    \end{array} \right.
\end{align}

\subsection{More Implementation Details}
During training, we reserve the dependency structures for low-frequency words even they are replaced with $\mathsf{unk}$. 
These $\mathsf{unks}$ are post-processed following the work in \cite{luong2014addressing}. 

Consider that the GCN and Graph Random Walks are basic modules in GNN area and have been well implemented in many PyTorch packages\footnote{https://www.dgl.ai/}, so our work is easy to modify these universal modules with dynamics to fit the sequence-to-sequence task.

We adopted Adam~\cite{kingma2014adam} as the optimizer with the momentum of 0.9 and the learning rate of $10^{-4}$ for updating parameters of our proposal. 
The mini-batch size is set to 96, and the decoding beam size is set to 12 for all the models. 

\subsection{About RIBES Criteria}
RIBES, considering more order information, is also used for Ja-En as suggest by \cite{isozaki2010automatic}.

As RIBES paper \cite{isozaki2010automatic} states, when we consider translation between distant language pairs such as Ja-En, some popular metrics (e.g. BLEU) don’t work well. Japanese and English have completely different sentence organizations, and special attention should be paid to word orders. E.g., some machine translation methods will translate “A because B” into “B because A”. Conventional metrics don’t significantly penalize such order mistakes. In our graph method, we will consider the syntactic graph in the model and pay more attention to the order information. 

\section{Experimental Details}

\subsection{Evaluated Benchmarks}
\label{app:dataset}

\noindent \textbf{NIST}. For the NIST OpenMT’s Chinese-English translation task, we leverage a subset of LDC corpus as bilingual data\footnote{LDC2003E14, LDC2005T10, LDC2005E83, LDC2006E26, LDC2003E07,
LDC2005T06, LDC2004T07, LDC2004T08, LDC2006E34, LDC2006E85, LDC2006E92, LDC2003E07, LDC2002E18, LDC2005T06}, which has around 2 M sentence pairs. For evaluation, the NIST2003 dataset is used as development data, and the test sets are NIST 2005, NIST 2006, NIST 2008, and NIST 2012. All English words are lowercase in both training and testing.

\noindent \textbf{WAT 2016}. In the Japanese-English and the English-Japanese translation tasks, the top 1 M sentence pairs from the ASPEC corpus\footnote{http://orchid.kuee.kyoto-u.ac.jp/ASPEC/} \cite{nakazawa2016aspec} is used for the Japanese-English task and 1.5 M sentence pairs
for the English-Japanese task. The development data contains 1,790 sentences, and the test data has 1,812 sentences.
We follow the pre-processing steps recommended in WAT 2016\footnote{http://lotus.kuee.kyoto-u.ac.jp/WAT/baseline/ \\ dataPreparationJE.html} as well as the official evaluation procedure provided by WAT 2016.


\noindent \textbf{WMT 2017}. For the English-German translation task, we use the pre-processed training data provided by the task organizers\footnote{http://data.statmt.org/wmt17/translation-task/preprocessed/de-en/}.
The training data includes around 5.8 million sentence pairs with 141 million English words and 135 million German words. We use the $\mathsf{newstest2013}$ as the validation set, and the $\mathsf{newstest2017}$ as the test set. 
We use 40 K sub-word tokens based on Byte Pair Encoding (BPE) \cite{sennrich2015neural} for both sides’ vocabularies.
For the Chinese-English translation task, we use the CWMT corpus, which consists of 9 M sentence pairs\footnote{http://www.statmt.org/wmt17/translation-task.html}. 
The $\mathsf{newsdev2017}$ is used as validation set and the $\mathsf{newstest2017}$ as the test set. 
For data pre-processing, we segment Chinese sentences with our in-house tool and segment English sentences with the scripts in Moses\footnote{https://github.com/moses-smt/mosesdecoder/blob/ \\ master/scripts/tokenizer
/tokenizer.perl}.
All other settings are the same as those for WMT 2017 English-German translation.

In addition, we also evaluate our method on another widely adopted benchmark namely \textbf{WMT 2014}.
For the English to German (En-De), the training set contains about 4.5 million parallel sentence pairs. 
We use the $\mathsf{newstest2013}$ and $\mathsf{newstest2014}$ as the validation and test sets, respectively.
The vocabulary is a 32K joint source and target byte pair encoding (BPE) \cite{sennrich2015neural}.
For evaluation, we adopt the case-sensitive BLEU scores by using the multi-bleu.perl script \footnote{https://github.com/mosessmt/mosesdecoder/blob/master  \\
/scripts/generic/multibleu.perl}.
For the English to French (En-Fr), the training set contains about 36 million parallel sentence pairs. 
We adopt the $\mathsf{newstest2013}$ and $\mathsf{newstest2014}$ as the validation and test sets, respectively. 
The 40K vocabulary is built based on a joint source and target BPE factorization.
The preprocessing and evaluation are the same as those for WMT’14 En-De.

\noindent \textbf{Parsing Tools}. 
For a fair comparison, we use the same dependency parsing tools of the compared methods.
For English and Chinese, we re-implement two arc-eager dependency parsers as in \cite{zhang2011transition} to generate the parse trees. 
The parsers are trained on the Penn Treebank and Chinese Treebank corpus. 
For Japanese, we use the J.DepP \cite{yoshinaga2014self} for parsing\footnote{http://www.tkl.iis.u-tokyo.ac.jp/ynaga/jdepp/}. For German, we use the Stanford Parser\footnote{https://nlp.stanford.edu/software/} \cite{manning2014stanford}. 

The parsing results are usually word level. 
We can accordingly modify the pseudo-golden dependency trees by a rule -- all subunits from one word are linked to the first unit with a new dependency label “subword”.

It is noted that besides syntactic graph generation, the proposed architecture can be generalized to any graph generative model, if some graph data can be well collected.

\subsection{Introduction of Compared Baselines}
\label{app:baselines}

We briefly introduce our compared baselines as follows:

\noindent \textbf{HPSMT}: An in-house reimplementation of the hierarchical phrase-based machine translation model \cite{chiang2005hierarchical}.
\noindent \textbf{SMT Hiero}: Hierarchical Phrase-based statistical machine translation (SMT) \cite{wu2018dependency}.
\noindent \textbf{SMT Phrase}: Phrase-based SMT \cite{wu2018dependency}.
\noindent \textbf{SMT S2T(T2S)}: String-to-Tree (Tree-to-String) Syntax-based SMT \cite{wu2018dependency}.
\noindent \textbf{RNNsearch}: An in-house reimplementation of the conventional RNN-based translation model \cite{bahdanau2014neural}.
\noindent \textbf{6-layer RNNsearch}: the RNNsearch model with a 6-layer decoder and a 6-layer encoder.
\noindent \textbf{Transformer}: the encoder-decoder framework released in \cite{vaswani2017attention}, where each decoder layer attends to the encoder output with multi-head attention.

Besides, we also compare our models with most existing syntax-aware translation methods that use source consistency/dependency trees or target consistency/dependency trees.
The representative methods are described as follows:

\noindent \textbf{ANMT}: \cite{eriguchi2016character} proposed an attention mechanism that enabled the decoder to generate a translated word while softly aligning it with phrases as well words of the source sentence.
\noindent \textbf{Tree2Seq}: \cite{chen2017improved} proposed a tree-to-sequence translation model by leveraging source constituency
trees\footnote{https://github.com/howardchenhd/Syntax-awared-NMT}.
\noindent \textbf{Bpe2tree}: \cite{aharoni2017towards} proposed a string-to-tree translation model named Bpe2tree. They replaced the target word sequence with the linearized constituency trees via the depth-first traversal order.
\noindent \textbf{Seq2Dep}: \cite{le2017improving} replaced target word sequences with linearized dependency trees via the depth-first traversal order.
\noindent \textbf{BiRNN+GCN}: \cite{bastings2017graph} used GCN to
produce syntax-aware word representations for source language. We adopt their most powerful model named
BiRNN+GCN for comparison.
\noindent \textbf{Semantic GCN}: \cite{marcheggiani2018exploiting} incorporated information about predicate-argument structure of source sentences (namely, semantic-role representations) into machine translation tasks.
We adopt their most powerful model with 4 GCN layers for comparison.
\noindent \textbf{SD(, SE+SD)-NMT}: \cite{wu2018dependency} proposed a Sequence-to-Dependency framework where the target translation and its corresponding dependence tree are jointly constructed and modeled during translation. In their work, the sequence-to-dependency neural machine translation model is denoted as SD-NMT and the dependency-to-dependency NMT model is denoted as SE+SD-NMT.

\subsection{Evaluation Results}
\label{app:app_exp}
\begin{table}
\centering
\begin{tabular}{l|c|c}
\toprule
\textbf{Method}            & \textbf{En-De} & \textbf{En-Fr} \\ \toprule
Transformer   & 27.3     & 41.0     \\ 
DBERT-NMT   & 27.5     & -    \\ 
MGNMT   & 27.7     & -     \\ 
CBBGCA   & \textbf{28.3}    & 41.5     \\ \bottomrule
\textbf{Dyn-STGCD}  & 28.1     & \textbf{41.8}     \\ \bottomrule
\end{tabular}
\caption{Evaluation results on the WMT 2014 English-German (En-De) and English-French (En-Fr) translation tasks with BLEU metric. The results of compared methods are taken from corresponding papers. Our BLEU is calculated in the same way described in main text. We DO NOT apply compound split for German post-processing and any model ensemble method such as checkpoints averaging.}
\label{tab7}
\end{table}

\noindent \textbf{Additional Quantitative Results.} 
Table \ref{tab5} gives the evaluation results on all test sets on the NIST Chinese-English task.
Overall, experimental results verify that our graph-based approach significantly improves translation qualities over the baseline NMT models and outperforms other syntax-based NMT methods.
In addition, we evaluate our approach with several leading methods including DBERT-NMT \cite{chen2019distilling}, MGNMT \cite{zheng2019mirror}, and CBBGCA \cite{zhou2022confidence} on WMT 2014.
As shown in Table \ref{tab7}, Dyn-STGCD achieves competitive results on both translation tasks.
It is noted that our implementations do not use any additional tricks which typically result in larger BLEU scores such as {\it average\_checkpoints} and {\it compound split}\footnote{https://github.com/facebookresearch/fairseq/tree \\ /main/examples/scaling\_nmt}.

\noindent \textbf{Predicting Syntactic Graphs.}
We select 500 instances from the NIST test sets to build an evaluation test set, where both source- and target-side do not contain $\mathsf{unk}$ and have a length between 15 and 35. 
We force our proposal to generate the reference and collect the corresponding syntactic graphs. 
As there are no golden graph references, directly evaluating parsing quality is not possible.
So we resort to estimating the consistency between the by-products of our approach and the parsing results from a stand-alone parser\footnote{https://github.com/explosion/spacy-models/releases}. 
The higher the consistency is, the closer the performance of the by-product is to the stand-alone parser. 
We parse the references to acquire pseudo graph references, then estimate the graphs generated by our method with the pseudo references. 


\section{Related Works}
\label{app:related}

\subsection{Structure-aware Machine Translation}
Structure information has long been argued as potentially useful for improving the performance of translation models.
 \citet{eriguchi2016tree} utilized syntactic information of the source sentence to guide the decoder output.
\citet{bastings2017graph} incorporated syntactic structure of the source sentence into machine translation using syntactic GCN.
\citet{wu2018dependency} introduced a sequence-to-dependency framework with the syntax-aware encoder and implicitly captured the dependencies of the target tokens, i.e., predicting the dependencies given generated target tokens. 
\cite{marcheggiani2018exploiting} is the first work to incorporate semantic structures into the encoder of translation models. 
Our work is mostly similar to the methodology in \cite{li2022structural} which focuses on leveraging the source structures to optimize word alignment but slightly harms the performance translation task. 
Our work differs from prior studies in that we focus on explicitly modeling the syntactic context of the target sentence to guide the translation, resulting in a novel decoder framework.

\subsection{Graph-based Text Generation}
GCN \cite{kipf2016semi} are specially designed to extract topological structures on graphs, which extend the concept of {\it convolution} to graph domain by designing operations that aggregate neighborhood information.
Previous studies verified that GCN can be applied to encode linguistic structures such as dependency trees \cite{bastings2017graph,marcheggiani2018exploiting,xu2018graph2seq,koncel2019text}. 
However, these works commonly use stacked and static GCN layers to capture the syntactic structure in the encoder side. 
Similar to the spirit of graph-to-graph model in molecular optimization task \cite{jin2018learning}, we exploit to dynamically model the syntactic dependencies of target tokens in the decoder but further use them to guide the translation task. 
To the best of our knowledge, this is the pioneering work that incorporates the graph-based syntactic constraints into the decoder of translation models.
We also argue that besides syntactic graph generation, our graph-to-graph architecture can be generalized to any graph generative model, if some graph data can be accurately collected.

\end{document}